\documentclass[twoside]{article}
\usepackage[accepted]{aistats2e}

\usepackage[]{graphicx} 
\usepackage{subfigure} 
\usepackage[numbers,square]{natbib}
\usepackage{algorithmicx}
\usepackage{algorithm}
\usepackage{algpseudocode}
\usepackage{amssymb,amsfonts,amsmath,amsbsy}
\usepackage{hyperref}
\usepackage{color}
\usepackage[usenames,dvipsnames]{xcolor}

\usepackage{listings}

\lstnewenvironment{bodycode}[2]{\small\lstset{caption=#1,label=#2,basicstyle=\small\ttfamily}}{}
\lstnewenvironment{code}[2]{\small\lstset{caption=#1,label=#2}}{}

\newenvironment{inline}[1]{{\small\ttfamily #1}}{}

\newcommand{\+}[1]{\ensuremath{\boldsymbol{#1}}}
\newcommand{\Normal}{\mathcal{N}}

\DeclareMathOperator*{\Var}{Var}

\algnewcommand\algorithmicyield{\textbf{yield~}}
\algnewcommand\Yield{\algorithmicyield}

\newcommand{\p}[2]{\ensuremath{p(#1 \,|\, #2)}}

\begin{document}

\twocolumn[
\aistatstitle{Tempering by Subsampling}
\aistatsauthor{Jan-Willem van de Meent \And Brooks Paige \And Frank Wood}
\aistatsaddress{Columbia University \And University of Oxford \And University of Oxford} ]

\begin{abstract}

In this paper we demonstrate that tempering Markov chain Monte Carlo samplers for Bayesian models by recursively subsampling observations without replacement can improve the performance of baseline samplers in terms of effective sample size per computation.  We present two tempering by subsampling algorithms, {\em subsampled parallel tempering} and {\em subsampled tempered transitions}.  We provide an asymptotic analysis of the computational cost of tempering by subsampling, verify that tempering by subsampling costs less than traditional tempering, and demonstrate both algorithms on Bayesian approaches to learning the mean of a high dimensional multivariate Normal and estimating Gaussian process hyperparameters.

\end{abstract}

\section{Introduction}

Markov chain Monte Carlo (MCMC) samplers for Bayesian models simulate a Markov chain whose equilibrium distribution is the posterior distribution $p(\theta|\+X) \propto p(\+X|\theta)p(\theta)$ of model parameters $\theta$ given a set of observations $\+X$ in a model composed of a likelihood  $p(\+X|\theta)$ and prior $p(\theta)$.    Tempered MCMC methods aim to improve baseline sampler mixing performance by simulating a Markov chain on an artificial joint distribution composed of auxilliary distributions at different temperatures.  Intuitively, traversing up and down the temperature ladder corresponds to heating and annealing.  
Tempering allows samplers to propose large jumps in good regions of the parameter space even for posterior distributions that are multi-modal or otherwise difficult to sample from, which in turn results in improved sampler mixing.
Tempered MCMC is closely related to simulated annealing, which samples  from successively more tightly peaked densities by descending an artificial temperature ladder once.  
Tempering methods are usually computationally costly.



In this paper we present an new approach to tempered MCMC for Bayesian models that reduces its computational cost. 
The idea behind our approach is exceedingly simple: a Markov chain can be heated by subsampling the data and then cooled again by adding the ``forgotten'' observations back in. 
Given that computational cost in most sampling procedures is dominated by the calculation of the likelihood or its gradients, tempering methods that use less data will be, by design, more computationally efficient than normal tempering methods. 
Like other tempering methods, tempering by subsampling is easy to implement as an outer loop that wraps an inner, general-purpose MCMC samplers including Metropolis \cite{Metropolis1953}, Metropolis Hastings (MH) \cite{Hastings1970}, Gibbs \cite{Geman:1984hh}, Hamiltonian Monte Carlo (HMC) \cite{duane1987hybrid,wang1987nonuniversal}, and Riemannian Monte Carlo (RMC) \cite{girolami2011riemann}.
The only requirement of the procedure is that the target density is conditioned on a set of observations that can be subsampled.

Tempering by subsampling must have higher computational cost per sample relative to the inner method that is employed.  However, if measured by effective sample size per computation, tempering by subsampling can actually be more efficient than its inner method if this increase in cost is offset by a larger increase in effective sample size. 
In two illustrative cases, a high dimensional Gaussian Bayesian mean estimation problem and a Gaussian process hyperparameter estimation problem, we have found that tempering by subsampling is more efficient than the baseline sampler in these terms.  It may be possible that this finding generalizes broadly; theoretical guarantees that it will seem unlikely.  Further empirical study seems warranted.   


The remainder of this paper is organized as follows. Sec.~\ref{sec:background} reviews tempering methods and techniques for estimating effective sample size from sampler output, Sec.~\ref{sec:termperingbysubsampling} covers tempering by subsampling  and includes asymptotic runtime analysis, Sec.~\ref{sec:experiments} contains experimental evidence of the computational and relative convergence characteristics of tempering by subsampling, and Sec.~\ref{sec:discussion} contains a discussion of our findings and suggestions for future work.




\section{Background}
\label{sec:background}


Tempering methods for MCMC use a series of auxilliary densities to interpolate between a target density and one that a Markov chain should be able to mix over rapidly \cite[]{geyer_chap_2011}.
In a Bayesian setting where one wishes to obtain samples from a posterior $h(\theta) = \p{\theta}{\+X} \propto p(\+X,\theta)p(\theta)$, a natural choice for such a set of interpolating densities is \cite{friel_jrssb_2008}
\begin{equation}
  \label{eq:tempering-bayes}
  h_m(\theta) 
  \propto 
  \exp[
    -\beta_m 
    \log \p{\+X}{\theta} 
    + 
    \log p(\theta)]
    ~.
\end{equation}
The degree of similarity between densities can be controlled by choosing $1 = \beta_0 > \beta_1 > \ldots > \beta_M = \beta_*$, which is often given a geometric form $\beta_m = \beta_*^{m/M}$. 
The $\beta$ parameter can be loosely interpreted as an inverse temperature. 
At $\beta_m = 1$ the unnormalized density $h_m(\theta)$ is the posterior. As $\beta_m \to 0$ the density converges to  the prior, from which it is often easy to sample. 

\subsection{Parallel Tempering}

Parallel tempering (PT) is an ensemble method that samples $\Theta = \{\theta_{m}\}$ jointly from $h(\Theta) = \prod_m h_m(\theta_m)$ at each iteration \cite[]{geyer_css_1991}.
Samples from synchronous parallel inner samplers running at each temperature are exchanged via swap proposals $\theta_n, \theta_m \to \theta_m, \theta_n$ that are accepted with probability 
\begin{equation}
    r(n,m) 
    =
    \min 
    \left[
    1
    ,
    \frac
    {h_n(\theta_m) h_m(\theta_n)}
    {h_n(\theta_n) h_m(\theta_m)}
    \right]
    ~.
\end{equation}
The swap move is its own inverse and leaves the joint distribution invariant.  Only samples from the low temperature marginal are collected. 

\subsection{Tempered Transitions}

The tempered transition (TT) method \cite[]{neal_sc_1996} is an MH procedure that uses a deterministic sequence of moves that raises the temperature to its highest level and then anneals it again.
During the ascending phase of the proposal a sample $\hat \theta_m$ is drawn from $h_m$ using a proposal density $q_{m}(\hat \theta'_{m-1}, \hat \theta_{m-1})$.
In the decending phase samples $\check \theta_m$ are drawn using proposals $q_{m}(\check \theta_m, \check \theta_{m+1})$. 
The entire trajectory therefore contains two values $\hat \theta_m$ and $\check \theta_m$ at each temperature, with the exception of $m=M$ where by convention we write $\hat \theta_m = \check \theta_m$.
The MH acceptance probability of the TT proposal arising from the full up and down traversal of the temperature ladder is 
\begin{equation}
    r(\check \theta_0, \hat \theta_0)
    =
    \min
    \left[
    1
    ,
    \prod_{m=1}^M
    \frac{h_{m}(\hat \theta_{m-1}) h_{m-1}(\check \theta_{m-1})}
         {h_{m}(\check \theta_{m-1}) h_{m-1}(\hat \theta_{m-1})}
    \right]
    .
\end{equation}
Note that the TT acceptance ratio can be interpreted as that of $M$ consecutive parallel tempering swaps.
On first inspection it may therefore appear that TT offers few advantages over PT methods, since the latter can obtain a new sample from the target density even when swap proposals are rejected, while offering similar mixing rates in expectation.
However the subtle difference between the two methods is that TT effectively evaluates swaps between pairs $\hat \theta_m, \check \theta_m$ that are both obtained using the same proposal mechanism, whereas PT evaluates swaps $\theta_m, \theta_n$ for samples obtained using two different mechanisms, allowing potentially higher acceptance rates in TT as compared to the product of swap acceptance rates in PT methods.

\subsection{Tempering in General}

When designing tempering methods there are a few things to choose. 
The first is the set of densities $h_m(\theta)$. 
Equation~\ref{eq:tempering-bayes} represents one possible scheme, another is to take $h_m(\theta) \propto p(\theta | \+X)^{\beta_m}$. 
In theory we may use any set of densities; practically they must be designed such that they are ``close.''  
The inner method for sampling at each temperature must also be chosen. 
A common choice is to use one or more MH steps, but HMC and other proposals may be used. 
Finally we must define a schedule for sampling and moving between temperatures.  

\subsection{Characterizing Sampler Performance}

A well-known property of MCMC samplers is that subsequent draws are often strongly correlated. 
For this reason, sampler performance is often characterized in terms of the effective sample size (ESS), i.e. the number of equivalent independent samples from the target density, which can be interpreted as a measure of the amount of information contained in a sampling chain.
The effective sample size can be defined in terms of an auto-correlation time $\tau$, the number of Markov chain transitions equivalent to a single independent draw.
Most commonly the autocorrelation time is estimated from a single simulation chain, using a batch means estimater, linear regression on the log spectrum and initial sequence estimates \cite{Thompson2010}.
The effective sample size is then obtained by dividing the number of sample draws by the autocorrelation time.

A deficicieny of single-chain estimators is that they tend to underestimate the autocorrelation time when a Markov chain has not fully converged to the equilibrium distribution.
This is particularly problematic when we wish to compare effective sample sizes obtained with tempering methods, since such methods are generally used in high-dimensional or multimodal cases where assessment of convergence is particularly difficult.
For this reason we characterize both MCMC convergence and effective number of samples using the estimated potential scale reduction $\hat R_\theta$ \cite{Gelman03}.
This quantity is calculated by running $C$ independent simulation chains, each from a different initialization.
After discarding the first half of our samples as a burn-in phase, we collect a total of $S$ samples from each chain.
For each parameter we obtain a sample estimate $\widehat{\Var}(\theta|\+X)$ of the marginal posterior variance $\Var(\theta|\+X)$
\begin{align}
    \widehat{\Var}(\theta|\+X) 
    &= 
    \frac{S-1}{S} W 
    + 
    \frac{1}{S} B
    .
\end{align}
The quantities $B$ and $W$ are known at the between-chain variance and within-chain variance respectively, which may be calculated as
\begin{align}
    B 
    &= 
    \frac{S}{C-1} 
    \sum_{c=1}^C(\bar \theta_c - \bar\theta)^2
    ,\\
    W 
    &= 
    \frac{1}{C} 
    \sum_{c=1}^C 
    \left[
        \frac{1}{S-1} 
        \sum_{s=1}^S 
        (\theta_{cs} - \bar\theta_{c}) 
    \right]
    ,
\end{align}
where $\theta_{cs}$ is sample $s$ from chain $c$. The sample estimate of the posterior variance is used to estimate the potential scale reduction
\begin{align}
    \label{eq:Rhat}
    \hat R_\theta 
    &= 
    \sqrt{{\widehat{\Var}(\theta|\+X)}/{W}}
    ,
\end{align}
which converges in expectation to $1$, from above, as the $C$ independent chains converge to the same distribution.
We follow the recommendation of \cite[p.~297]{Gelman03}, and consider our sampler to have mixed adequately when $\hat R_\theta < 1.1$.
Finally, the total effective sample size across all chains can be calculatedfrom the sample estimate of the posterior variance as
\begin{align}
    \label{eq:ess}
    ESS_\theta 
    &= 
    C \times S \times \: 
    \min
    \left[ 1, \widehat{\Var}(\theta|\+X) / B \right]
    .
\end{align}
When characterizing the computational performance per unit computation we report the effective sample size, normalized by the wall clock computation time, averaged over chains.

\section{Tempering by Subsampling}
\label{sec:termperingbysubsampling}


\begin{algorithm}[!t]
    \caption{\label{alg:spt} Subsampled Parallel Tempering}
    \begin{algorithmic}
        \State $\+X_0 \gets \+X,N_0 \gets |\+X|, s \gets 1$ 
        \State $\{\beta_m\} \gets$ initialize inverse ``temperatures''
        \State $\{\theta_{m,0}\} \gets$ initialize all chain starting values
        \For{$m = 1 \ldots M$}
            \State $N_m \gets$ \inline{round(}$\beta_m |X_0|$\inline{)}
            \State $\+X_m \gets$ \inline{sample-without-replacement(}$\+X_{m+1}, N_m$\inline{)}
        \EndFor
        \For{$s=1\ldots S$} 
            \For{$m = 1 \ldots M$}
                \State $\theta_{m,s} \gets$ \inline{transition}$(\theta_{m,s-1} | \+X_m)$
            \EndFor
            \For{$m = M \ldots 1$}
                \State $\rho' \gets h_{m}(\theta_{m-1,s}) h_{m-1}(\theta_{m,s}) $
                \State $\rho \gets h_{m}(\theta_{m,s}) h_{m-1}(\theta_{m-1,s})$
                \If{$\rho'/\rho > $ \inline{rand()}}
                    \State $\theta_{m,s}, \theta_{m+1,s} \gets \theta_{m+1,s}, \theta_{m,s}$
                \EndIf
            \EndFor
            \State $s \gets s+1$
        \EndFor \\
        \Yield $\{\theta_{0,s}\}$
    \end{algorithmic}
\end{algorithm}

Subsampled variants of tempering methods function exactly like normal tempering methods, with the exception that auxilliary distributions take the form
\begin{equation}
    h_m(\theta) \propto \p{\+X_m}{\theta} p(\theta)
    ~,
\end{equation}
where each $\+X_m$ is a subsample of size \mbox{$N_m \simeq \beta_m N$} of the full data $\+X$.
We can choose to either recursively subsample $\+X_{m+1} \subset \+X_{m}$ (without replacement), or independently sample $N_m$ observations at each temperature (also without replacement).
We do not present a detailed comparison of these two strategies here.
We did however perform simple trials that indicated that independent subsamples can lead to very low swap acceptance rates when $N_m \ll N$, even when using a small spacing in the temperature ladder.
For this reason we here employ recursive subsamples for the purposes of our experiments.

Subsampled parallel tempering (SPT) method can be devised by picking a set of recursive subsamples $\+X_m$ during initialization that are then the remainder of the sampling procedure (see Algorithm \ref{alg:spt}). 
In the implementation used here, we propose a series of moves with $n=m-1$, starting at $m=M$ and moving down systematically until $m=1$.
In this manner, a sample obtained at any temperature can be accepted in the target density with some probability at each iteration.

The subsampled tempered transitions (STT) variant (see Algorithm \ref{alg:stt}) recursively subsamples the observations at each jump $m \to m+1$ in the upward temperature sweep. 
The attractive feature of this scheme is that $\+X_{m+1} \subset \+X_m$, while avoiding the subsample bias of the SPT method, since different subsamples are chosen for each sampler iteration.

\begin{algorithm}[!t]
    \caption{Subsampled Tempered Transitions}
    \label{alg:stt}
    \begin{algorithmic}
        \State $\+X_0 \gets \+X,N_0 \gets |\+X|, s \gets 1$ 
        \State $\{\beta_m\} \gets$ initialize inverse ``temperatures''
        \State $\theta_{0,0} \gets$ initialize chain 

        \For{$s=1\ldots S$} 
            \State $\hat{\theta}_0 \gets \theta_{0,s-1}$
            \For{$m = 1 \ldots M$}
              \State $N_m \gets$ \inline{round(}$\beta_m |X_0|$\inline{)}
            \State $\+X_m \gets$ \inline{sample-without-rep$\ldots$(}$\+X_{m-1}, N_m$\inline{)}
                \State $\hat{\theta}_{m} \gets$ \inline{transition(}$\hat{\theta}_{m-1} | \+X_m$\inline{)}
                \State $\hat{\rho}_{m} \gets h_m(\hat{\theta}_{m-1}) / h_{m-1}(\hat{\theta}_{m-1})$
            \EndFor
            \State $\check{\theta}_M \gets \hat{\theta}_M$
            \For{$m = M\!-\!1 \ldots 0$}
                \State $\check{\theta}_m \gets$ \inline{transition(}$ \check{\theta}_{m+1} | \+X_m$\inline{)}

                \State $\check \rho_{m+1} \gets h_m(\check{\theta}_{m}) / h_{m+1}(\check{\theta}_{m})$
            \EndFor
            \If{$(\sum_{m=1}^{M} (\hat \rho_m + \check \rho_m)) > $ \inline{rand()}}
                \State $\theta_{0,s} \gets \check{\theta}_0$
            \Else
                \State $\theta_{0,s} \gets \theta_{0,s-1}$
            \EndIf
        \EndFor \\ 
        \Yield $\{\theta_{0,s}\}$
    \end{algorithmic}
\end{algorithm}

\subsection{Computational Complexity}

\label{sec:complexity}

To assess the computational cost of subsampled tempering variants relative to their non-subsampling counterparts, we will assume that sampling from the target distribution requires computation time $\tau_0 = \tau_{*} N^{\alpha}$, where $\alpha \ge 1$ is some exponent that depends on the type of model and proposal mechanism, and $\tau_*$ is a constant pre-factor. The time per sample for $m>0$ is then simply given by
\begin{equation}
    \tau_m 
    = \tau_{*} N_m^{~\alpha} 
    = \tau_{*} (\beta_m N_0)^\alpha
    = \tau_0 \beta_m^\alpha
    .
\end{equation}
 Let $S = \sum S_m$ denote the total number of samples in in the chain, where $S_m = s_m S_0$ represents the number of samples from each $h_m(\theta)$. 
 The total computation time needed to draw $S$ samples is 
\begin{align}
  T
  &= 
  \sum_{m=1}^MS_m \tau_m 
  = 
  \sum_{m=1}^M(S_0 s_m) (\tau_0 \beta_m^{~\alpha})
  \,,
  \\
  &= 
  S_0 \tau_0 \sum_{m=1}^Ms_m \beta_m^{~\alpha}
  \,.
\end{align}
The computational complexity of FT sampling can therefore be controlled to some extent by the choice of inverse temperature ladder $\beta_m$, which is generally given an exponential form $\beta_m = \beta_*^{m/M}$.
In SPT an equal amount of samples are generated at each temperature level, i.e. $s_m = 1$,  and the form above is a geometric series whose sum evaluates to
\begin{align}
  T^{\textsc{\tiny SPT}}
  =
  S_0 \tau_0
  \frac{1 - \beta_{*}^{\alpha (1 + 1/M)}}
       {1 - \beta_{*}^{\alpha / M}}
  \,.
\end{align}
In TT methods we must sample from each distribution twice per iteration, with the exception of $h_0$ and $h_M$, where we obtain only one sample. 
For this case we can express the computation time as
\begin{align}
  T^{\textsc{\tiny STT}}
  &=
  2 T^{\textsc{\tiny SPT}}
  - 
  S_0 \tau_0
  \left(
    1 + \beta_*^{\alpha / M}
  \right)
  \,.
\end{align}
If we normalize by the time $T^0 = S_0 \tau_0$ required to draw the same number of samples using a non-tempering algorithm, we obtain the ratios
\begin{align}
    \tau^{\textsc{\tiny SPT}}
    &=
    \frac{1 - \beta_{*}^{\alpha (1 + 1/ M)}}
         {1 - \beta_{*}^{\alpha / M}}
    \,,
   \label{eq:tau_SPT}
    \\
    \tau^{\textsc{\tiny STT}}
    &=
    \left[
    2
    \frac{1 - \beta_{*}^{\alpha (1 + 1/M)}}
         {1 - \beta_{*}^{\alpha / M}}
    -
    (1 + \beta_*^{\alpha / M})
    \right]
    \,.
    \label{eq:tau_STT}
\end{align}
By comparison, the relative cost for non-subsampling tempering variants is
\begin{align}
    \tau^{\textsc{\tiny PT}}
    &= 
    M 
    \,,
    \\
    \tau^{\textsc{\tiny TT}}
    &= 
    2(M - 1)
    \,.
\end{align}
In short, the factors influencing the relative computational cost are the number of temperature levels $M$, the smallest inverse temperature $\beta_*$, and the exponent $\alpha$ that determines the asymptotic scaling with the number of observations of the computation.
Using a larger number of temperature levels increases the acceptance rate of swap proposals in PT variants, and the base acceptance rate in TT methods.
In subsampling variants, lowering $\beta_*$ reduces the computational cost, albeit at the expense of decreasing the acceptance rates.
The quantities $\beta_*$ and $M$ need to to be adjusted to match the difficulty of the inference problem. 
Depending on the values chosen, subsampling reduces the computational cost of tempering by a factor 2 to 10.

A caveat to the analysis presented here is that the evaluation of swap proposals is in general more expensive in subsampling approaches, which require two extra evaluations of the likelihoods $\log \p{x_n}{\theta_m}$ and  $\log \p{x_m}{\theta_n}$.
When the base sampling method is cheap (e.g. a single MH step), evaluation of the swap proposals can represent a significant fraction of the computational cost.
We should therefore expect subsampling to be most effective when the base sampling mechanism is expensive, as will generally be the case when multiple MH steps or HMC sampling are used.



\section{Experiments}
\label{sec:experiments}

\begin{figure*}[!t]
  \includegraphics{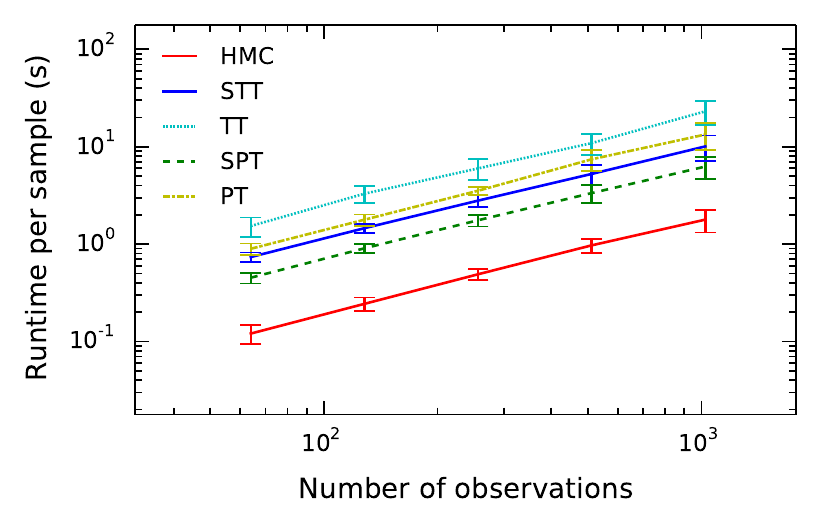}
  \includegraphics{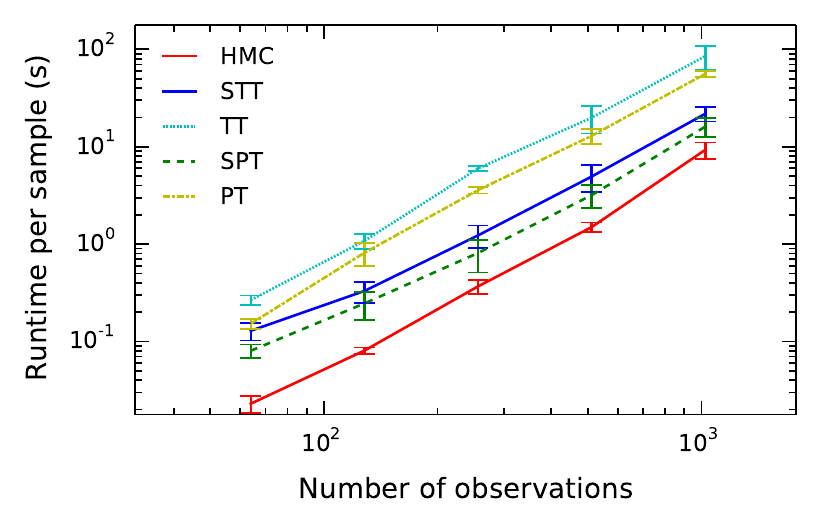}
  \caption{Run time as a function of dataset size. (left) In the multivariate normal model run time is $\mathcal{O}(N)$. 
  The observed computational cost relative to HMC is 
 3.51 (SPT),  5.65 (STT), 7.48 (PT),  and 12.94 (TT).
  (right) In the GP regression model the run time is governed by the cost of the Cholesky decomposition of an $N\times N$ kernel matrix.
At $N=1024$ observations, the cost relative to HMC is 1.73 (SPT), 2.35 (STT), 6.03 (PT), and 9.19 (TT).
}
\label{fig:runtime}
\end{figure*}

We test tempering by subsampling on two problems: sampling the posterior mean in a model with a multivariate Gaussian likelihood and sampling the hyperparameters in a Gaussian Process regression problem. We find that tempering by subsampling is in general advantageous.

In each model we are interested in characterizing how the performance of tempering methods, as measured in terms of effective sample size per unit computation, depends on the number of observations and the dimensionality of the parameter space. 
To this end we perform two sets of sweeps, one with fixed dimensionality and varying dataset size, and one varying dimensionality and fixed dataset size.

In order to more reliably determine the autocorrelation time and effective sampling size in each experiment, we run 3 different chains, which are initialized at $\theta = {\theta_*, \theta_*/2, 2\theta_*}$, where $\theta_* = E_{p(\theta)}[\theta]$ is the expected value of the parameters under the prior.
We assess convergence using the estimated potential scale reduction $\hat R$ (Eq. \ref{eq:Rhat}), assuming convergence when the median value of $\hat R$ over all dimensions drops below $1.1$. 
Similarly we asses the effective sample size ESS$_{\theta}$ (Eq.~\ref{eq:ess}) in terms of the median across dimensions.
The criteria are less strict than using the minimum value across dimensions, as generally recommended in the statistics literature \cite[p.~297]{Gelman03}, but are less sensitive to outliers in high-dimensional problems. 


\subsection{Multivariate Gaussian}

In this set of experiments we assume $N$ observations $\+X$ are distributed according to a $D$-dimensional multivariate normal with unknown mean $\theta$ and known covariance $\Sigma$, with a prior on $\theta$ of the same form
\begin{align}
    \p{\+X}{\theta} 
    &= 
    \prod_n^N \mathcal{N}(\+x_n \,|\, \theta, \Sigma)
    ,
    \\
    p(\theta) 
    &= 
    \prod_d^D \mathcal{N}(\theta_d \,|\, 0, \sigma_0^2)
\end{align}
We now simulate a set of observations from this generative model, and sample the posterior mean.
This posterior is of course easy to calculate analytically and sample from directly. It is included as a diagnostic to assess sampler performance in problems where the posterior is smooth and unimodal, but may have correlated variables.

In these experiments we characterize dependence on the dimensionality in a set of runs where $N=256$, and $D=5,10,50,100$. The dependence on dataset size is evaluated at $D=50$, and $N=64,128,256,512,1024,2048$. 
In each set of experiments we use a single random seed, implying that the first $N=64$ observations in the $N=128$ experiment will be identical to those in the $N=64$ experiment, for any given model.

\begin{figure*}[!t]
\includegraphics{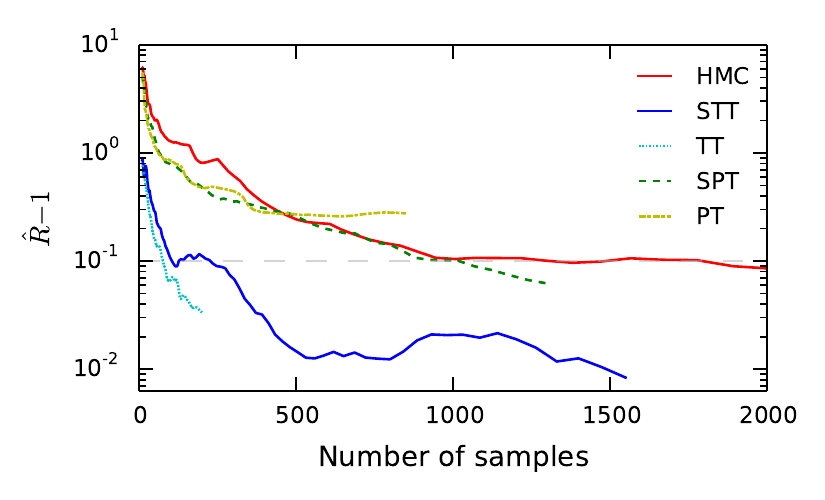}
  \includegraphics{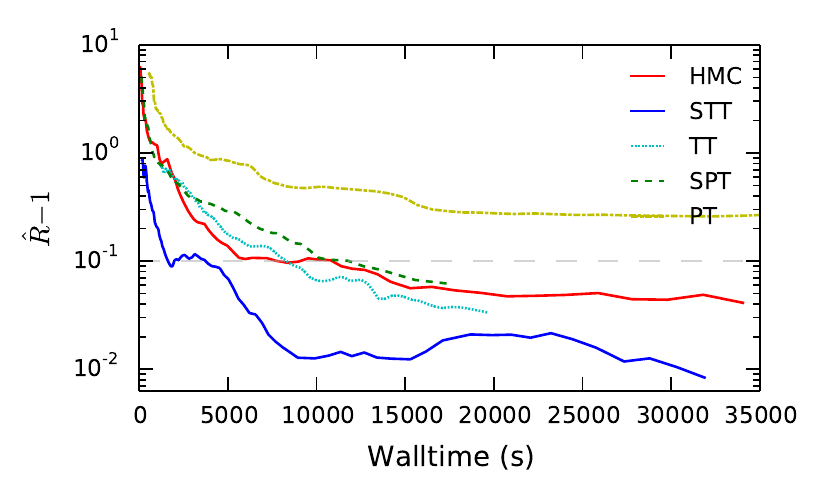}
  \caption{Convergence of MCMC sampling in GP regression with $N=512$ and $|\theta|=20$, as a function of the number of samples (left) and wall time (right). Lines show the median estimated potential scale reduction, with the dashed line marking the threshold $\hat R = 1.1$.
 Both STT and TT converge much faster per sample, and per unit computation STT outperforms all other algorithms.
  }
  \label{fig:convergence}
\end{figure*}

\subsection{Gaussian Process Regression}

To characterize performance in models where the cost of the likelihood scales in a non-linear manner with the number of observations, we sample the hyperparameters in Gaussian Process (GP) regression problems \cite{Rasmussen06}.
We include this problem as a test-case for performance of tempering by subsampling methods in cases where the base sampling procedure has a computational cost and log-likelihood that scales in a non-linear fashion with the number of data points.


In Gaussian Process regression, we use a squared exponential automatic relevance determination (ARD) kernel
\begin{align}
    \kappa(\+x, \+x') 
    &= 
    \sigma_f^2 
    \exp \bigg\{ 
        - \sum_{d=1}^D 
          \frac{(x_d - x_d')^2}
               {2 \ell_d^2} 
    \bigg \}
    ~,
\end{align}
parameterized by $D$ characteristic length scale parameters $\ell_1, \dots \ell_D$ and a vertical scale parameter $\sigma_f$.
Our observations $\+y$ have a Gaussian likelihood
\begin{align}
    \label{eq:gen-y}
    \+y \sim \Normal( \+0, \+K + \sigma_n^2\+I_N )
\end{align}
where $\sigma_n^2$ is an observation noise variance, $\+I_N$ is the $N\times N$ identity matrix, and we have defined the matrix $\+K$ such that each element $K_{ij}  = \kappa(\+x_i, \+x_j)$.
Each of the $D+2$ parameters $\theta = \{ \ell_1, \dots, \ell_D, \sigma_f, \sigma_n \}$ are constrained to be non-negative.
We impose a log-Gaussian prior on each length scale, and gamma priors on the vertical scale and noise terms
\begin{align}
    \ell_d &\sim \ln \Normal(\mu_0, \sigma_0) \\
    \sigma_f &\sim {\Gamma}(a_f, b_f) \\
    \sigma_n &\sim {\Gamma}(a_n, b_n)
\end{align}
where
\begin{align}
    \ln \Normal(x|\mu,\sigma) 
    &= 
    \frac{1}{x\sqrt{2\pi}\sigma} 
    \exp 
    \bigg\{ 
        -\frac{(\ln x - \mu)^2}
              {2\sigma^2} 
    \bigg\}
\end{align}
and
\begin{align}
    {\Gamma}(x|a,b) &= \frac{b^a}{\Gamma(a)} x^{a-1} e^{-bx}
\end{align}

In our GP regression runs we draw a set of hyperparameters $\theta$ from the prior and simulate $N$ observations by first drawing a set coordinates $\+x_1, \dots, \+x_N$ and then sampling the corresponding observations $\+y_1, \ldots, \+y_N$ from a GP according to Eq.~\ref{eq:gen-y}.
We then  run tempering MCMC to sample $\theta \sim \p{\theta}{\+Y}$.

In all experiments, the hyperparameters are set to $\mu_0 = 0.5$, $\sigma_0 = 1$, $a_f = 4$, $b_f = 1$, $a_n = 2$, $b_n = 2$.
The coordinates $\+X$ are sampled from a $D$-dimensional isotropic Gaussian centered at $\+0$, with standard deviation $\sum_d \ell_d / D$. 
Our sweeps are run at fixed dimension $D=18$, with $N=64,128,256,512,1024$ and fixed $N=512$ with $D=3,8,13,18$.

\begin{figure*}
  \includegraphics[width=0.33\textwidth]{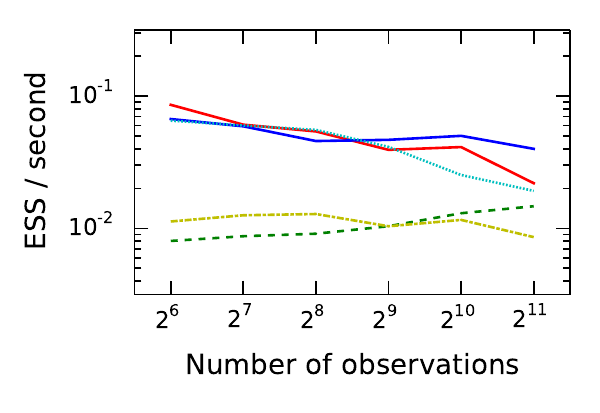}
  \includegraphics[width=0.33\textwidth]{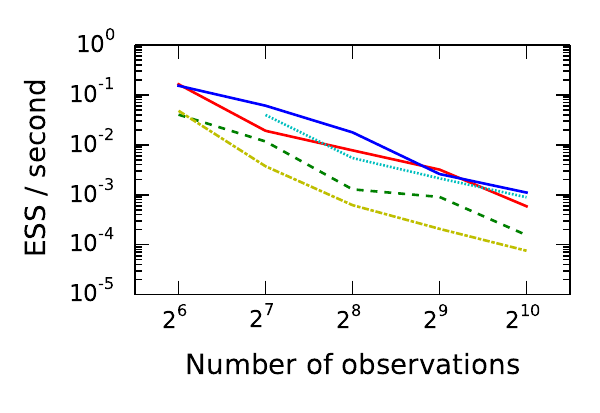}
  \includegraphics[width=0.33\textwidth]{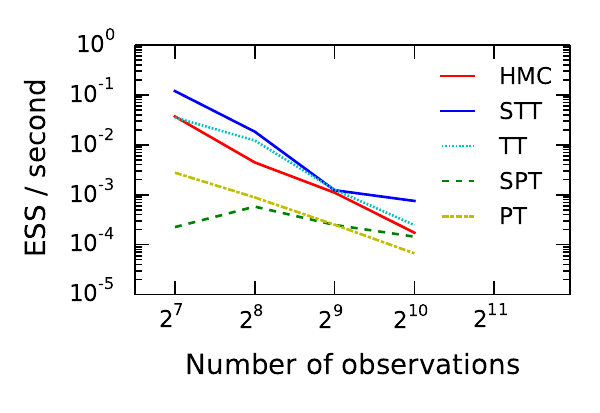}
  \caption{Comparison of effective samples per second over varying numbers of observations $N$, for a fixed dimension $|\theta|$. 
  (left) Multivariate Normal with HMC, $|\theta| =50$;
  (center) Gaussian Process with MH, $|\theta| = 20$;
  (right) Gaussian Process with HMC, $|\theta| = 15$.
  We see that the relative benefit of all tempering methods increases with $N$; in particular, STT consistently outperforms both other tempering methods and the inner sampler as $N$ becomes sufficiently large.
  }
  \label{fig:varies_with_N}
\end{figure*}

\subsection{Sampling Procedure}

All experiments are run using SPT and STT samplers, the corresponding non-subsampling variants PT and TT, and a basic sampler where no tempering is performed.
We employ $M = 6$ auxiliary distributions, using a geometric spacing of the inverse temperatures $\beta_m = 2^{-m/2}$, resulting in a minimum value $\beta_* = 1/8$.

Tempering can be performed using any base procedure that draw samples $\theta \sim h_m(\theta)$. 
In these experiments we use two types of inner sampling procedures. The first is a simple one-step Metropolis-Hastings updates. The second is Hamiltonian Monte Carlo (HMC) sampling \cite{neal_chap_2010}, which offers state-of-the-art performance in the multivariate normal and GP regression models.

Metropolis-Hastings updates use a standard Gaussian proposal $\theta' \sim \mathcal{N}(\theta \mid \sigma^2)$ where we follow \cite{neal_sc_1996} and use a step size proportional to the temperature $\sigma_m = 0.1 \beta_m^{-1/2}$, which results in approximately uniform acceptance rates at all temperature levels.

The HMC proposal mechanism performs numerical integration of a trajectory in the parameter space, which requires specification of the step size $\epsilon$ and a number steps $L$.
Automatic tuning of these two parameters is an area of active research \cite{hoffman_arxiv_2011,girolami2011riemann}. 
Here we are primarily interested in comparing performance between non-tempered, tempered, and tempering by subsampling methods. 
We therefore set both parameters to fixed values for each set of experiments, using $\epsilon = 0.01$ in both experiments, with $L=10$ for the multivariate normal and $L=5$ for the GP regression models.

In models where parameters values are constrained, it is often convenient to integrate HMC trajectories in a transformed, unconstrained, set of coordinates. In GP regression we use coordinates $\log \theta$ in the case of GP regression, where $\theta > 0$ for all parameters.

%

\subsection{Run time analysis}

Empirical run times for the Gaussian likelihood and GP regression models can be seen in Fig.~\ref{fig:runtime}. 
The per-sample complexity of the inner procedure is $\mathcal{O}(N)$ in the first, and order $\mathcal{O}(N^3)$ for the second, since likelihood and gradient evaluation require a Cholesky decomposition of an $N\times N$ kernel matrix in this model.

As expected, SPT and STT require less computation per sample than their traditional counterparts. 
Based on the analysis in Section \ref{sec:complexity}, we can use Eqs.~\ref{eq:tau_SPT}--\ref{eq:tau_STT} for $\tau^\text{SPT}$ and $\tau^\text{STT}$ to estimate the additional computational cost factors associated with subsampling tempering, relative to the inner sampler.
For the multivariate Gaussian model, $\alpha = 1$; at $M = 6$ and $\beta_* = 1/8$ we expect the additional computational cost to be a factor of $3.11$ for SPT and a factor of $4.52$ for STT, as $N \rightarrow \infty$.
In the Gaussian Process example, $\alpha = 3$; in this case the additional computational cost of tempering is asymptotically only a factor of $1.55$ for SPT and $1.74$ for STT.

In the multivariate normal model, the observed computational cost relative to HMC is 3.51 (SPT), 5.65 (STT), 7.48 (PT),  and 12.94 (TT).
In the GP regression model, the cost relative to HMC is 1.73 (SPT), 2.35 (STT), 6.03 (PT), and 9.19 (TT).
Up to small differences these values are in good agreement with the asymptotic analysis presented in Section \ref{sec:complexity}.




\subsection{Convergence rate}

One measure of relative performance is the time it takes for the sampling distribution to converge to the target density. 
We track the estimated scale reduction factor $\hat R$ as a function of wall clock time, and of the number of samples drawn.
In Fig.~\ref{fig:convergence} we show the median value of $\hat R$, computed over each dimension of $\theta$, for a GP model with $N=512$ and $|\theta|=D+2=20$.
We observe that the TT and STT methods converge an order of magnitude faster than non-tempered HMC, both in terms of sample count and computation wall time.

\begin{table}[!t]
\begin{tabular}{l|ccccc}
$|\theta|$ & HMC & STT & TT & SPT & PT \\ \hline
$5$ & $0.1378$ & $\mathbf{0.1828}$ & $0.0445$ & $0.0682$ & $0.0189$ \\
$10$ & $0.0027$ & $\mathbf{0.0053}$ & $0.0022$ & $0.0011$ & $0.0002^\star$ \\
$15$ & $0.0011$ & $0.0012$ & $\mathbf{0.0013}$ & $0.0002^\star$ & $0.0003$ \\
$20$ & $0.0007$ & $\mathbf{0.0033}$ & $0.0016$ & $0.0009$ & $0.0001^\star$ \\
$25$ & $0.0010$ & $\mathbf{0.0016}$ & $0.0011$ & $0.0003^\star$ & $0.0003^\star$
\\ 
\hline
\hline
$|\theta|$ & MH & STT & TT & SPT & PT \\ \hline
$5$ & $0.0279$ & $\mathbf{0.0503}$ & $0.0495$ & $0.0262$ & $0.0204$ \\
$10$ & $0.0028$ & $\mathbf{0.0151}$ & $0.0065$ & $0.0005$ & $0.0003$ \\
$15$ & $0.0021$ & $\mathbf{0.0038}$ & $0.0035$ & $0.0005$ & $0.0002$ \\
$20$ & $\mathbf{0.0032}$ & $0.0026$ & $0.0021$ & $0.0009$ & $0.0002$
\end{tabular}
\caption{ Median effective samples per second when sampling hyperparameters of a Gaussian Process with $N = 512$ observations,
as evaluated over a range of models with different parameter dimensionality $|\theta|$.
The most computationally efficient sampler for each is marked in bold;
entries where all three independent chains have not fully converged to $\hat R < 1.1$ are marked with a star.
Results are shown using
(top) an HMC inner sampler; 
(bottom) an MH inner sampler.
Per unit computation, STT is consistently the best performing sampling algorithm.
}
\label{table:gp_observations}
\end{table}

\begin{table}[!t]
\begin{tabular}{l|ccccc}
$|\theta|$ & HMC & STT & TT & SPT & PT \\ \hline
$5$ & $\mathbf{0.7915}$ & $0.1380$ & $0.0681$ & $0.2201$ & $0.1089$ \\
$10$ & $\mathbf{0.3334}$ & $0.1361$ & $0.0652$ & $0.2172$ & $0.1119$ \\
$50$ & $\mathbf{0.0537}$ & $0.0363$ & $0.0257$ & $0.0142$ & $0.0091$ \\
$100$ & $0.0101$ & $\mathbf{0.0145}$ & $0.0101$ & $0.0026$ & $0.0020$
\\
\hline
\hline
%
$|\theta|$ & MH & STT & TT & SPT & PT \\ \hline
$5$ & $0.9949$ & $0.4903$ & $\mathbf{3.1858}$ & $0.0344$ & $0.1665$ \\
$10$ & $0.1445$ & $0.2664$ & $\mathbf{0.7232}$ & $0.0148$ & $0.0390$ \\
$50$ & $0.0221$ & $\mathbf{0.0579}$ & $0.0524$ & $0.0012$ & $0.0018$ \\
$100$ & $0.0102$ & $\mathbf{0.0240}$ & $0.0221$ & $0.0008^\star$ & $0.0010^\star$
\end{tabular}
\caption{ Median effective samples per second in multivariate Gaussian models
with dimensionality 5, 10, 50, and 100.
Results are shown for
(top) $N=1024$ observations, HMC inner sampler; (bottom) $N=256$ observations, MH inner sampler.
}
\label{table:mvn_observations}
\end{table}

\subsection{Dataset Size Dependence}

The overall computational cost of drawing a sample grows as the size of the dataset increases.
In Fig.~\ref{fig:varies_with_N}
we exhibit the effective number of samples drawn per second, across a number of data sizes, 
for two different GP models, each with a different inner sampler,
and for a $50$-dimensional multivariate Gaussian model.
In general, we see that subsampled tempered transitions outperforms all other approaches
on the GP models, across all dimensions.
In the simpler multivariate Gaussian model, we see the tempering methods become more competitive as the dataset becomes larger.

\subsection{Parameter Dimensionality Dependence}

We also investigate the relative performance of tempering schemes across parameter dimensionality $|\theta|$,
for both models and for both MH and HMC inner samplers.
Experimental results showing the performance of these sampling schemes are presented 
for both models, with both MH and HMC inner samplers, in 
Table~\ref{table:gp_observations} and Table~\ref{table:mvn_observations}.
We see that overall sample efficiency decreases across all models as $|\theta|$ increases;
in the Gaussian Process model, STT consistently outperforms other methods per unit computation, across dimensionality. 
In the simpler multivariate Gaussian model, tempering becomes more effective as the dimensionality increases.

\section{Discussion}
\label{sec:discussion}

Tempering methods are typically employed only when other sampling approaches perform suboptimally \cite{geyer_chap_2011}.
While the field of physical chemistry has adopted tempered MCMC on the basis of empirical successes \cite{marinari_epl_1992,hansmann_cpl_1997,trebst_jcp_2006,chodera_jctc_2007,prinz_jcp_2011}, usage by the machine learning community has been relatively limited \cite{salakhutdinov_nips_2009,salakhutdinov_aistats_2010,desjardins_aistats_2010}.

Subsampling approaches to tempering MCMC might change that.  
Obvious practical limitations prevent us from exhaustively characterizing the computational effect of tempering by subsampling for all possible models and datasets; 
however, our initial results are encouraging. 
Once a base sampling procedure is in place, subsampled tempering methods require very little additional code, so practitioners should be able to test the effectiveness of these methods in their applications with relative ease.

We also wish to note that our implementation of subsampled parallel tempering may be suboptimal in the sense that it subsamples the data once and then retains the resulting subsets for the duration of the sampling procedure. 
We believe that it might be possible to construct a valid Markov chain wherein the  subsamples are resampled on some interval.  In that case it may also be possible to obtain an unbiased estimator of $\beta_m \log \p{\+X}{\theta} + \log p(\theta)$ \cite{beaumont_gen_2003,andrieu_as_2009}, allowing subsampled parallel tempering to be used to obtain estimates of Bayes factors \cite{calderhead_csda_2009}.

A practical caveat to keep in mind when characterizing the performance of tempering methods is that calculating the ESS per computation time is by no means a trivial exercise.  Most commonly, ESS is computed via an autocorrelation time obtained from single-chain estimators such as batch means, linear regression on the log spectrum and initial sequence estimates \cite{Thompson2010}. 
The problem with these in-chain methods is that they underestimate the autocorrelation time when a sampler has yet to converge.
Consequently a simple comparison of autocorrelation times may underrepresent the effectiveness of tempering when the base sampling method does not fully converge.   We have taken great care to run all samplers until convergence criteria, use the best estimator of ESS we know of, and to conduct computation time measurements appropriately.

\newpage

\bibliographystyle{plainnat}
\bibliography{bibtex/jwm_machine_learning,bibtex/brooks_additions}

\begin{thebibliography}{26}
\providecommand{\natexlab}[1]{#1}
\providecommand{\url}[1]{\texttt{#1}}
\expandafter\ifx\csname urlstyle\endcsname\relax
  \providecommand{\doi}[1]{doi: #1}\else
  \providecommand{\doi}{doi: \begingroup \urlstyle{rm}\Url}\fi

\bibitem[Andrieu and Roberts(2009)]{andrieu_as_2009}
Christophe Andrieu and Gareth~O. Roberts.
\newblock {The pseudo-marginal approach for efficient Monte Carlo
  computations}.
\newblock \emph{The Annals of Statistics}, 37\penalty0 (2):\penalty0 697--725,
  April 2009.

\bibitem[Beaumont(2003)]{beaumont_gen_2003}
Mark~a Beaumont.
\newblock {Estimation of population growth or decline in genetically monitored
  populations.}
\newblock \emph{Genetics}, 164\penalty0 (3):\penalty0 1139--60, July 2003.

\bibitem[Calderhead and Girolami(2009)]{calderhead_csda_2009}
Ben Calderhead and Mark Girolami.
\newblock {Estimating Bayes factors via thermodynamic integration and
  population MCMC}.
\newblock \emph{Computational Statistics \& Data Analysis}, 53\penalty0
  (12):\penalty0 4028--4045, October 2009.

\bibitem[Chodera et~al.(2007)Chodera, Swope, Pitera, Seok, and
  Dill]{chodera_jctc_2007}
John~D. Chodera, William~C. Swope, Jed~W. Pitera, Chaok Seok, and Ken~a. Dill.
\newblock {Use of the Weighted Histogram Analysis Method for the Analysis of
  Simulated and Parallel Tempering Simulations}.
\newblock \emph{Journal of Chemical Theory and Computation}, 3\penalty0
  (1):\penalty0 26--41, January 2007.

\bibitem[Desjardins et~al.(2010)Desjardins, Courville, Bengio, Vincent, and
  Delalleau]{desjardins_aistats_2010}
Guillaume Desjardins, Aaron Courville, Yoshua Bengio, Pascal Vincent, and
  Olivier Delalleau.
\newblock {Parallel Tempering for Training of Restricted Boltzmann Machines}.
\newblock In \emph{AISTATS}, volume~9, pages 145--152. IEEE, July 2010.

\bibitem[Duane et~al.(1987)Duane, Kennedy, Pendleton, and
  Roweth]{duane1987hybrid}
Simon Duane, Anthony~D Kennedy, Brian~J Pendleton, and Duncan Roweth.
\newblock Hybrid {M}onte {C}arlo.
\newblock \emph{Physics letters B}, 195\penalty0 (2):\penalty0 216--222, 1987.

\bibitem[Friel and Pettitt(2008)]{friel_jrssb_2008}
N.~Friel and a.~N. Pettitt.
\newblock {Marginal likelihood estimation via power posteriors}.
\newblock \emph{Journal of the Royal Statistical Society}, 70\penalty0
  (3):\penalty0 589--607, July 2008.

\bibitem[Gelman et~al.(2003)Gelman, Carlin, Stern, and Rubin]{Gelman03}
Andrew Gelman, John~B. Carlin, Hal~S. Stern, and Donald~B. Rubin.
\newblock \emph{Bayesian Data Analysis}.
\newblock Chapman and Hall/CRC, 2003.
\newblock ISBN 158488388X.

\bibitem[Geman and Geman(1984)]{Geman:1984hh}
S.~Geman and D.~Geman.
\newblock {Stochastic Relaxation, {G}ibbs Distributions, and the {B}ayesian
  Restoration of Images}.
\newblock \emph{IEEE TPAMI}, \penalty0 (6):\penalty0 721--741, November 1984.

\bibitem[Geyer()]{geyer_chap_2011}
Charles~J Geyer.
\newblock In Steve Brooks, Andrew Gelman, Galin Jones, and Xiao-Li Meng,
  editors, \emph{Handbook of Markov Chain Monte Carlo}, chapter~11, pages
  295--311. Chapman and Hall/CRC, May .

\bibitem[Geyer(1991)]{geyer_css_1991}
CJ~Geyer.
\newblock {Markov Chain Monte Carlo Maximum Likelihood}.
\newblock In \emph{Computing Science and Statistics, Proceedings of the 23rd
  Symposium on the Interface}, volume~5, pages 156--63. Interface Foundation,
  January 1991.

\bibitem[Girolami and Calderhead(2011)]{girolami2011riemann}
Mark Girolami and Ben Calderhead.
\newblock Riemann manifold {L}angevin and {H}amiltonian {M}onte {C}arlo
  methods.
\newblock \emph{Journal of the Royal Statistical Society: Series B (Statistical
  Methodology)}, 73\penalty0 (2):\penalty0 123--214, 2011.

\bibitem[Hansmann(1997)]{hansmann_cpl_1997}
Ulrich~H.E. Hansmann.
\newblock {Parallel tempering algorithm for conformational studies of
  biological molecules}.
\newblock \emph{Chemical Physics Letters}, 281\penalty0 (1-3):\penalty0
  140--150, December 1997.

\bibitem[Hastings(1970)]{Hastings1970}
W.~K. Hastings.
\newblock Monte {C}arlo sampling methods using {M}arkov chains and their
  applications.
\newblock \emph{Biometrika}, 57(1):\penalty0 97--109, 1970.

\bibitem[Hoffman and Gelman(2011)]{hoffman_arxiv_2011}
Matthew~D Hoffman and Andrew Gelman.
\newblock {The No-U-Turn Sampler: Adaptively Setting Path Lengths in
  Hamiltonian Monte Carlo}.
\newblock \emph{arXiv}, \penalty0 (2008):\penalty0 30, November 2011.

\bibitem[Marinari and Parisi(1992)]{marinari_epl_1992}
E~Marinari and G~Parisi.
\newblock {Simulated Tempering : A New Monte Carlo Scheme}.
\newblock \emph{EPL (Europhysics Letters)}, 451, 1992.

\bibitem[Metropolis et~al.(1953)Metropolis, Rosenbluth, Rosenbluth, Teller, and
  Teller]{Metropolis1953}
A.~W. Metropolis, A.~W. Rosenbluth, M.~N. Rosenbluth, A.~H. Teller, and
  E.~Teller.
\newblock Equations of state calculations by fast computing machines.
\newblock \emph{Journal of Chemical Physics}, 21:\penalty0 1087--1092, 1953.

\bibitem[Neal(1996)]{neal_sc_1996}
Radford~M. Neal.
\newblock {Sampling from multimodal distributions using tempered transitions}.
\newblock \emph{Statistics and Computing}, 6\penalty0 (4):\penalty0 353--366,
  December 1996.

\bibitem[Neal(2010)]{neal_chap_2010}
RM~Neal.
\newblock {MCMC using Hamiltonian dynamics}.
\newblock In \emph{Handbook of Markov Chain Monte Carlo}, pages 113--162. 2010.

\bibitem[Prinz et~al.(2011)Prinz, Chodera, Pande, Swope, Smith, and
  No\'{e}]{prinz_jcp_2011}
Jan-Hendrik Prinz, John~D Chodera, Vijay~S Pande, William~C Swope, Jeremy~C
  Smith, and Frank No\'{e}.
\newblock {Optimal use of data in parallel tempering simulations for the
  construction of discrete-state Markov models of biomolecular dynamics.}
\newblock \emph{The Journal of chemical physics}, 134\penalty0 (24):\penalty0
  244108, June 2011.

\bibitem[Rasmussen and Williams(2005)]{Rasmussen06}
Carl~Edward Rasmussen and Christopher K.~I. Williams.
\newblock \emph{Gaussian Processes for Machine Learning (Adaptive Computation
  and Machine Learning)}.
\newblock The MIT Press, 2005.
\newblock ISBN 026218253X.

\bibitem[Salakhutdinov(2009)]{salakhutdinov_nips_2009}
Ruslan Salakhutdinov.
\newblock {Learning in Markov random fields using tempered transitions}.
\newblock \emph{Advances in neural information \ldots}, pages 1--9, 2009.

\bibitem[Salakhutdinov(2010)]{salakhutdinov_aistats_2010}
Ruslan Salakhutdinov.
\newblock {Learning deep Boltzmann machines using adaptive MCMC}.
\newblock \emph{Proceedings of the 27th International Conference on Machine
  Learning}, 1, 2010.

\bibitem[{Thompson}(2010)]{Thompson2010}
M.~B. {Thompson}.
\newblock {A Comparison of Methods for Computing Autocorrelation Time}.
\newblock \emph{ArXiv e-prints}, October 2010.

\bibitem[Trebst et~al.(2006)Trebst, Troyer, and Hansmann]{trebst_jcp_2006}
Simon Trebst, Matthias Troyer, and Ulrich H~E Hansmann.
\newblock {Optimized parallel tempering simulations of proteins}.
\newblock \emph{The Journal of chemical physics}, 124\penalty0 (17):\penalty0
  174903, May 2006.

\bibitem[Wang and Swendsen(1987)]{wang1987nonuniversal}
JianÑSheng Wang and RH~Swendsen.
\newblock Nonuniversal critical dynamics in {M}onte {C}arlo simulations.
\newblock \emph{Physical review letters}, 1987.

\end{thebibliography}

\end{document}